%% file: samplepaper.tex
\documentclass[runningheads]{llncs}

\usepackage[T1]{fontenc}
\usepackage{graphicx}
\usepackage{booktabs}
\usepackage{array}
\usepackage{placeins}
\usepackage{amsmath,amssymb}
\usepackage{marvosym}
\usepackage{tikz}
\usetikzlibrary{arrows.meta}
\begin{document}

\title{Replan, Repair, or Edit? A Unified Empirical Evaluation of Travel Agents for Itinerary Revision under Resource Disruptions}
\titlerunning{Travel Agents for Itinerary Revision}
\author{
Xiaofei Yuan\inst{1} \and
Yan Zhang\inst{1} \and
Shaobo Qiao\inst{1} \and
Huangleshuai He\inst{2} \and
Leyan Ni\inst{2} \and
Mingchen Ju\inst{1} \and
Lujia Yang\inst{3} \and
Sijia Xu\inst{2} \and
Yifu Tang\inst{4} \and
Zhengyi Yang\inst{3}\textsuperscript{(\Letter)}
}

\authorrunning{X. Yuan et al.}

\institute{
\textsuperscript{1} Euler AI;
\textsuperscript{2} University of New South Wales;
\textsuperscript{3} University of Sydney;
\textsuperscript{4} Vecton AI\\
% \textsuperscript{1} Euler AI, Australia;
% \textsuperscript{2} University of New South Wales, Sydney, Australia;
% \textsuperscript{3} University of Sydney, Sydney, Australia;
% \textsuperscript{4} Vecton AI, Australia\\
\email{\{xiaofei.yuan,yan.zhang,shaobo.qiao,mingchen.ju\}@eulerai.au};
\email{\{huangleshuai.he,leyan.ni,sijia.xu\}@unsw.edu.au};
\email{\{lyan8923,zhengyi.yang\}@sydney.edu.au};
\email{yves.tang@vectonai.com}
}
% \title{Replan, Repair, or Edit? A Unified Empirical Evaluation of Travel Agents for Itinerary Revision under Resource Disruptions}
% \titlerunning{Travel Itinerary Revision under Resource Disruptions}
% \author{
% Xiaofei Yuan\inst{1} \and
% Yan Zhang\inst{1} \and
% Shaobo Qiao\inst{1} \and
% Huangleshuai He\inst{2} \and
% Leyan Ni\inst{2} \and
% Mingchen Ju\inst{1} \and
% Lujia Yang\inst{3} \and
% Sijia Xu\inst{2} \and
% Yifu Tang\inst{4} \and
% Zhengyi Yang\inst{3}
% }

% \authorrunning{X. Yuan et al.}

% \institute{
% Euler AI, Australia\\
% \email{\{xiaofei.yuan,yan.zhang, shaobo.qiao, mingchen.ju\}@eulerai.au}\\
% % \email{Zhang.11022.osu@gmail.com}
% \and
% University of New South Wales, Sydney, Australia\\
% \email{\{huangleshuai.he, leyan.ni, sijia.xu\}@unsw.edu.au}\\
% % \email{leyan.ni@student.unsw.edu.au}
% \and
% University of Sydney, Sydney, Australia\\
% \email{\{lyan8923, zhengyi.yang\}@sydney.edu.au}
% \and
% Vecton AI, Australia\\
% \email{yves.tang@vectonai.com}
% }

\maketitle

\begin{abstract}
Travel-planning agents generate itineraries that may become infeasible after acceptance because of flight cancellations, hotel unavailability, or attraction closures. Revising these itineraries involves full replanning, classical plan repair, and LLM-based travel-agent revision, whose differing task formulations and evaluation protocols hinder comparison. We conduct a systematic empirical study using two TREK-derived benchmark sets: 500 single-disruption cases, including feasible and infeasible instances, and 200 feasible simultaneous compound-disruption cases. We compare LLM--Z3 full replanning, IPyHOPPER hierarchical repair, and an iTIMO local-revision adapter across effectiveness, plan stability, and computational cost. LLM--Z3 with Gemini achieved the highest observed compound-disruption success. IPyHOPPER nearly matched that configuration's single-disruption overall success, while preserving substantially more of the accepted itinerary on successful repairs. Successful hierarchical and local repairs made fewer edits and retained more accepted commitments than full replanning. Computational profiles differed: IPyHOPPER used no LLM inference, the evaluated LLM--Z3 adapter used compact one-call inference, and the iTIMO adapter consumed substantially more tokens. The study provides practical guidelines for balancing feasibility recovery, commitment preservation, and computational cost within evaluated settings.
\keywords{Travel planning \and Itinerary revision \and Plan repair \and Language agents \and Empirical evaluation}
\end{abstract}
\input{main}

\bibliographystyle{splncs04}
\bibliography{mingchen_refs,lujia_refs,leyan_refs,new}

\end{document}

%% file: main.tex
\section{Introduction}
\label{sec:introduction}

Travel itinerary planning coordinates transportation, accommodation, and activities under user preferences, budgets, schedules, and resource availability. Travel-planning agents can generate feasible itineraries, including by combining large language models (LLMs) with formal verification tools \cite{zhang2025,xie2024travelplanner,hao2025formalverify,zhao2026trace,jiang2026nsmem}. Flight cancellations, hotel unavailability, or attraction closures can subsequently invalidate an accepted itinerary \cite{karmakar2026triptide,oh2025flex}, shifting the task from \emph{initial generation} to \emph{revision}.

\begin{figure}[t]
  \centering
  \input{fig_problem}
  \caption{Itinerary revision under resource disruptions, resulting in either a feasible revision or an infeasibility diagnosis.}
  \label{fig:problem}
\end{figure}
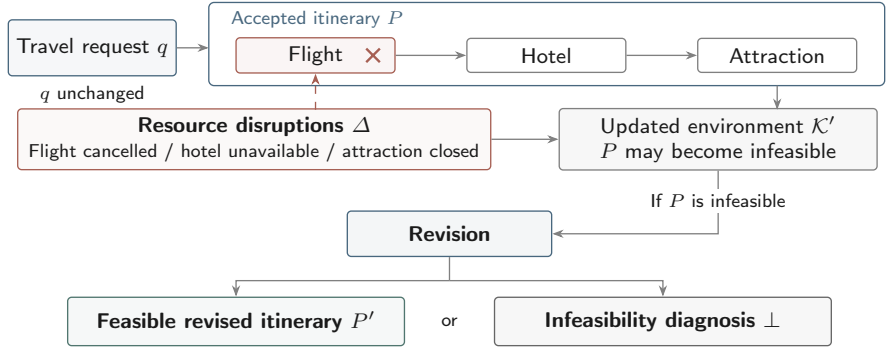

Revision starts from accepted commitments and preferences. Restoring feasibility may require changes beyond the unavailable resource, but generating a new itinerary can also replace arrangements that remain valid. Evaluation must therefore consider preservation of the accepted itinerary~\cite{fox2006planstability} alongside feasibility recovery and computational cost, as illustrated in Figure~\ref{fig:problem}. Studies of LLM evaluation and agentic systems across classification, question answering, retrieval, financial analysis, data exploration, and serving provide complementary examples of task-specific quality, reliability, and efficiency assessment~\cite{wu2024experimental,lai2025graphy,yang2024parallel,tang2025tabular,liu2026a2rag,shu2026forexagent,yang2026raids,liu2026hypersu,huan2025scaling}.

Three paradigms differ in their use of the accepted itinerary: \emph{full replanning} solves the updated problem anew, \emph{classical plan repair} reuses plan structure, and \emph{LLM-based travel-agent revision} edits the itinerary directly \cite{hao2025formalverify,zaidins2025htnrepair,huang2026itimo}. We compare these scopes by effectiveness, preservation, and computational cost.

\noindent\textbf{Motivation.}
Four research gaps motivate the comparison. \textbf{(1) Limited benchmark coverage for itinerary revision.} Travel-planning benchmarks emphasize itinerary generation and broader planning capabilities \cite{xie2024travelplanner,qi2026trek,chaudhuri2025tripcraft,chen2026traveleval,cheng2026travelbench}. Despite work on adaptive planning and itinerary modification \cite{karmakar2026triptide,huang2026itimo}, benchmark support for revising accepted itineraries under resource disruptions remains limited, particularly for evaluating both single and compound disruptions. \textbf{(2) Lack of a unified evaluation setting.} Full replanning, classical plan repair, and LLM-based travel-agent revision originate from different task formulations and evaluation settings, so previously reported results are not directly comparable. A common revision interface, feasibility criterion, and controlled disruption setting are needed to compare them. \textbf{(3) Incomplete evaluation dimensions.} Success rates do not reveal how much of an accepted itinerary is preserved or at what computational cost. High repair coverage may come with extensive changes, while local revision may incur high inference or runtime costs despite making few edits. \textbf{(4) Limited strategy-selection guidance.} Existing evaluations provide limited guidance for choosing a revision paradigm when deployments prioritize feasibility recovery, plan preservation, and computational cost.

\noindent\textbf{Contributions.}
We compare LLM--Z3 full replanning, IPyHOPPER hierarchical repair, and an iTIMO-based local revision adapter, with four contributions:
\begin{itemize}
    \item \textbf{A benchmark for itinerary revision under resource disruptions.} Two fixed TREK-derived benchmark sets \cite{qi2026trek} provide 500 single-disruption cases, including feasible and infeasible instances, and 200 feasible compound-disruption cases. All methods receive the same instances.
    \item \textbf{A unified evaluation setting for heterogeneous revision strategies.} A common interface takes the original request, accepted itinerary, disruptions, and updated environment, and assesses complete repairs or infeasibility outcomes against a common feasibility criterion. Unsuccessful attempts count as failures; method-specific restrictions and budgets are explicit.
    \item \textbf{A multi-dimensional evaluation framework.} Effectiveness covers feasible repair, overall success, and correct refusal; stability covers logical edits and preservation; computational cost covers latency, LLM calls, and tokens.
    \item \textbf{Practical guidelines for revision-strategy selection.} Observed trade-offs characterize the strengths, limitations, and suitable deployment settings of the three paradigms for the evaluated implementations and settings.
\end{itemize}

\section{Background and Problem Formulation}
\label{sec:background}

This section formalizes the travel-itinerary revision problem considered in this study. We define the planning environment, resource disruptions, revision outcomes, plan stability, and the feasibility-first repair objective.

\subsection{Travel Itinerary Planning}
\label{sec:plan-generation}

A travel request $q$ specifies the user's requirements and preferences for a trip, such as destinations, dates, and budget. The planning environment $\mathcal{K}$ contains knowledge-base records, resource availability, and execution conditions needed to assess executability. An itinerary $X$ is a complete plan comprising flights or other transportation between destinations, accommodation, local transportation where applicable, and attractions or activities with temporal and spatial assignments.

Let $C(q,\mathcal{K})$ be the set of applicable constraints. Request constraints require the itinerary to satisfy the trip requirements in $q$. Grounding requires its resources and associated information to be supported by the planning environment. Persona or preference constraints capture the traveller's specified needs, while budget constraints bound the cost of the trip. Temporal consistency concerns the timing and compatibility of scheduled components; spatial consistency concerns their locations and the connections between them. These constraints determine the feasible set
\begin{equation}
\mathcal{F}(q,\mathcal{K})=\{X\mid c(X)=\mathrm{true}\ \text{for every }c\in C(q,\mathcal{K})\}.
\label{eq:feasible-itineraries}
\end{equation}
Initial itinerary generation seeks any complete $X\in\mathcal{F}(q,\mathcal{K})$. Once accepted, the itinerary is denoted by $P$, with $P\in\mathcal{F}(q,\mathcal{K})$ before disruption.

\subsection{Resource Disruptions and Revision Outcomes}
\label{sec:problem-formulation}

A resource is a travel entity used in an itinerary; this study considers disruptions to flights, hotels, and attractions. A disruption $\delta$ makes a previously available resource unavailable. Let $\Delta=\{\delta_1,\ldots,\delta_k\}$ denote the set of disruptions: $|\Delta|=1$ is a \emph{single disruption}, and $|\Delta|>1$ is a \emph{compound disruption}. Compound disruptions in this study are presented simultaneously, rather than arriving as a sequence of online changes. They produce the updated environment
\[
\mathcal{K}'=\operatorname{Update}(\mathcal{K},\Delta).
\]

The request $q$ remains unchanged, but disruptions may render $P$ infeasible in $\mathcal{K}'$ through temporal and spatial dependencies. For example, a replacement flight may shift the arrival time and prevent a scheduled attraction visit.

The common revision interface takes the original request, accepted itinerary, disruptions, and updated environment:
\begin{equation}
f(q,P,\Delta,\mathcal{K}')\longrightarrow P',\ \bot,\ \text{or}\ \mathrm{fail}.
\label{eq:revision-interface}
\end{equation}
Here $P'$ denotes a complete revised itinerary; it is a successful repair only if $P'\in\mathcal{F}(q,\mathcal{K}')$. The outcome $\bot$ denotes a correct infeasibility diagnosis, requiring $\mathcal{F}(q,\mathcal{K}')=\varnothing$. The outcome $\mathrm{fail}$ records an attempt that produces neither a valid repair nor a justified infeasibility outcome within the method's execution procedure or budget. Examples include timeouts, execution errors, exhausted search or correction budgets, and malformed or invalid candidates.

Failure to find a feasible repair does not establish infeasibility. The original user requirements are not silently relaxed to make the updated problem feasible.

\subsection{Plan Stability}
\label{sec:stability}

Feasible revisions can differ in how much of $P$ they retain. Plan stability is assessed through \emph{logical itinerary components}, or \emph{logical slots}, each representing a semantically meaningful commitment rather than a raw JSON field or textual string. Examples include a flight assigned to a route, a hotel booking for a city, and an attraction assignment for a day.

The logical edit measure $d(P,P')$ counts semantic itinerary changes, not textual edits. Let $S(P)$ denote the logical slots of $P$. Preservation is the fraction of slots in $S(P)$ whose assignments remain unchanged in $P'$. Fewer edits and higher preservation generally indicate greater stability. The measures are complementary rather than mathematically equivalent. Section~\ref{sec:repair-metrics} specifies the benchmark's matching and counting rules and reports stability on successful feasible repairs.

\subsection{Feasibility-First Repair Objective}
\label{sec:objective}

Feasibility is the primary requirement: retaining accepted commitments does not compensate for an invalid itinerary. When $\mathcal{F}(q,\mathcal{K}')$ is nonempty, an ideal minimum-edit revision satisfies
\begin{equation}
P^*\in\underset{X\in\mathcal{F}(q,\mathcal{K}')}{\arg\min}\ d(P,X).
\label{eq:repair-preference}
\end{equation}
This conceptual preference favors fewer changes among feasible alternatives; it is neither an objective implemented by all methods nor a guarantee of globally minimum-edit repair.

\section{Related Work and Representative Strategy Selection}
\label{sec:strategies}

Existing work relevant to itinerary revision can be organized into three broad lines according to how an accepted plan is used after the environment changes. First, travel-planning and solver-assisted approaches typically formulate the updated request as a new planning problem and generate a complete feasible itinerary, with limited emphasis on preserving the previously accepted plan \cite{xie2024travelplanner,hao2025formalverify}. Second, classical plan-repair methods explicitly reuse an existing plan, retaining prior assignments or structural decompositions while revising the parts affected by changed goals or execution conditions \cite{fox2006planstability,zaidins2025htnrepair}. Third, recent LLM-based itinerary-modification methods operate directly on the current itinerary through semantic editing operations rather than reconstructing a formal plan \cite{huang2026itimo}. These lines differ mainly in their use of the accepted itinerary and the scope of revision. We therefore group the literature into \emph{full replanning}, \emph{classical plan repair}, and \emph{LLM-based travel-agent revision}, and evaluate one representative per category.

\subsection{Full Replanning: LLM--Z3}
\label{sec:full-replan}
Full replanning seeks a complete feasible itinerary for an updated problem without necessarily preserving $P$. TravelPlanner \cite{xie2024travelplanner} and AgentTravel \cite{zhao2025agenttravel} address itinerary generation and knowledge-augmented planning. External reasoning supports constraint satisfaction through verification in LLM-Modulo \cite{gundawar2024llmmodulo}, mixed-integer optimization in To the Globe \cite{ju2024totheglobe}, formalized programming in LLMFP \cite{hao2025llmfp}, and LLM--solver integration in Personal Travel Solver \cite{shao2025pts}. RETAIL \cite{deng2025retail} and ATLAS \cite{choi2026atlas} address complex or changing requirements without centering accepted-itinerary preservation.

Hao et al.~\cite{hao2025formalverify} translate requests into executable steps and code invoking a Satisfiability Modulo Theories (SMT) solver. Formal solving separates constraint satisfaction from language generation and supports full replanning without a preservation objective.

Our \emph{LLM--Z3 Full Replan} adapter uses a fixed Python/Z3 model \cite{demoura2008z3} rather than generated solver code. One LLM call extracts structured requirements; the model solves updated resource constraints and decodes $P'$. It does not optimize similarity to $P$, which is retained for post-hoc stability evaluation.

\subsection{Classical Plan Repair: IPyHOPPER}
\label{sec:ipyhopper}
Classical repair reuses plans under changed goals, constraints, or execution conditions. Related work spans assignment reuse in dynamic constraint satisfaction \cite{verfaillie1994solutionreuse}, localized plan adaptation \cite{gerevini2000planadaptation}, planning-based repair \cite{vanderkrogt2005}, and plan stability and minimum-change repair \cite{fox2006planstability,saetti2025}. Hierarchical Task Network (HTN) planning decomposes tasks into subtasks and primitive actions, providing structure for repair methods including SHOP-FIXER, IPyHOPPER, and REWRITE \cite{zaidins2025htnrepair}.

IPyHOPPER \cite{zaidins2025htnrepair} reuses plan structure while adapting repair scope. Given decomposition tree $T$ and primitive sequence $\pi=\operatorname{plan}(T)$, simulation identifies a failed action and replaces an ancestor task's decomposition. If the result is inapplicable, backtracking expands repair to a higher-level task. This reuse does not guarantee minimum edits.

Our adapter reconstructs $T$ from $P$ before activating $\Delta$. Updated conditions identify the affected action, and the repaired primitive sequence is decoded into $P'$. Candidates target disrupted slots; the worker uses domain constraints without the benchmark scorer. Simulation is internal and does not imply that the traveller has executed a trip prefix.

\subsection{LLM-Based Travel-Agent Revision: iTIMO}
\label{sec:itimo}
LLM-based revision edits an itinerary semantically rather than reconstructing a formal plan or repairing an HTN hierarchy. iTIMO \cite{huang2026itimo} provides an operation-based formulation over points of interest (POIs):
\begin{equation}
\mathcal{O}=\{o_{\mathrm{add}},o_{\mathrm{replace}},o_{\mathrm{delete}}\}.
\label{eq:itimo-operations}
\end{equation}
These operations add, replace, or delete a POI. The source benchmark evaluates POI-level perturbations based on popularity, spatial distance, and category diversity. iTIMO's local modification task motivates our editing baseline; its formulation is not a general resource-disruption repair algorithm.

Our iTIMO adapter receives $q$, $P$, $\Delta$, and $\mathcal{K}'$ and selects operations and targets, leaving untouched components unchanged. Unlike the source's single POI-level modification, it permits corrections for simultaneous disruptions within the budget in Section~\ref{sec:protocol}. Corrections modify the preceding result and do not represent new arrivals of $\Delta$; the adapter remains a bounded local-repair baseline.

\begin{table}[!htbp]
  \centering
  \footnotesize
  \setlength{\tabcolsep}{2pt}
  \caption{Conceptual comparison of the evaluated revision strategies.}
  \label{tab:strategy-comparison}
  \begin{tabular}{@{}>{\raggedright\arraybackslash}p{0.15\textwidth}>{\raggedright\arraybackslash}p{0.18\textwidth}>{\raggedright\arraybackslash}p{0.24\textwidth}>{\raggedright\arraybackslash}p{0.16\textwidth}>{\raggedright\arraybackslash}p{0.20\textwidth}@{}}
    \toprule
    Paradigm & Representative & Role of accepted $P$ & Repair scope & Core mechanism \\
    \midrule
    Full replanning & LLM--Z3 adapter & Used only for stability evaluation & Global & Formal constraint solving \\
    Classical plan repair & IPyHOPPER & Decomposition from $P$ reused & Adaptive hierarchical & HTN repair and backtracking \\
    LLM-based revision & iTIMO adapter & $P$ directly edited & Bounded local & ADD, REPLACE, DELETE \\
    \bottomrule
  \end{tabular}
\end{table}

\subsection{Comparison of Representative Strategies}
\label{sec:strategy-comparison}

Table~\ref{tab:strategy-comparison} highlights the main differences among the three representative strategies. Full replanning treats the updated problem globally and does not use the accepted itinerary $P$ as a repair structure, whereas IPyHOPPER explicitly reuses the decomposition derived from $P$ and adapts the repair scope hierarchically. The iTIMO adapter instead edits $P$ directly through bounded local operations. These differences yield three distinct revision scopes---global, adaptive hierarchical, and bounded local---and motivate their comparative evaluation under a common revision setting.

\section{Experimental Evaluation}
\label{sec:experiments}

\begin{figure}[t]
  \centering
  \input{fig_evaluation}
  \caption{Unified evaluation of the three representative revision strategies across effectiveness, plan stability, and computational cost.}
  \label{fig:evaluation}
\end{figure}
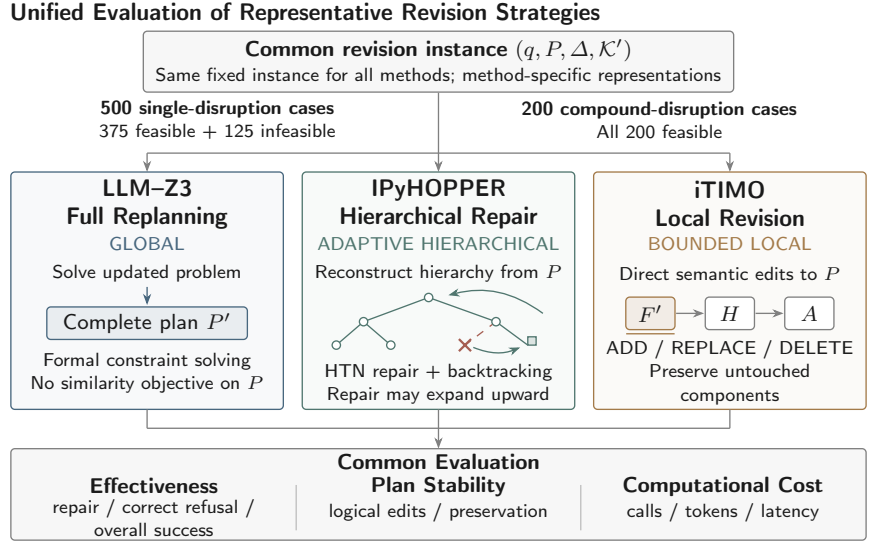

Figure~\ref{fig:evaluation} summarizes the unified evaluation setting. We evaluate the three representative strategies using identical fixed instances under both single and simultaneous compound resource disruptions. The evaluation covers \emph{effectiveness}, \emph{plan stability}, and computational cost, capturing repair coverage, preservation of accepted commitments, and execution or inference overhead.

\subsection{Experimental Setup}
\label{sec:dataset}
\label{sec:protocol}
The experimental setup specifies the benchmark instances, evaluation metrics, and implementation settings used for all compared strategies.

\noindent\textbf{Benchmark and tasks.}
All methods receive identical fixed TREK-derived instances \cite{qi2026trek}: a request, accepted itinerary, and disruptions in an updated environment. Reference repairs and feasibility labels are hidden.

The 500 single-disruption cases comprise 375 feasible flight, optional-hotel, and optional-attraction disruptions and 125 infeasible required-hotel conflicts, testing repair and infeasibility detection under unchanged requirements. The 200 feasible compound cases comprise 50 each of AB, AC, BC, and ABC (A/B/C: flight/hotel/attraction disruptions), with multiple components invalidated simultaneously. The benchmarks share 191 source queries but are reported separately; compound groups use different itineraries without difficulty matching.

\noindent\textbf{Evaluation metrics.}
\label{sec:repair-metrics}
We evaluate each configuration along three complementary dimensions: effectiveness, plan stability, and efficiency. Together, these metrics capture whether a revision succeeds, how much of the accepted itinerary it preserves, and the computational cost required to obtain it.

\noindent\emph{Effectiveness.} Valid repairs are complete itineraries passing applicable TREK checks for requirements, grounding, persona, budget, and spatio-temporal consistency, and avoiding disrupted resources. Single-disruption \emph{feasible repair rate}, \emph{correct refusal rate}, and \emph{overall success rate} are percentages: valid repairs divided by 375, correct refusals divided by 125, and their sum divided by 500, respectively. For the 125 required-hotel conflicts, correct refusal checks recognition of the constructed conflict, not a general-purpose proof of infeasibility. Errors, invalid outputs, and timeouts in retained final records count as failures in the denominators, not correct refusals. All compound cases are feasible: overall success equals repair rate over 200 cases; refusal is inapplicable.

\noindent\emph{Plan stability.} Over successful feasible repairs, \emph{mean logical edits} averages $d(P,P')$; \emph{preservation rate} averages the percentage of original logical slots retaining the same resource or value at the corresponding key. Flights are keyed by route, hotels and cars by city, and city assignments and attraction sets by day. Daily attraction edits count as $\max(|\mathrm{removed}|,|\mathrm{added}|)$. These implement Section~\ref{sec:stability}'s semantic measures. Table rows use each configuration's successful feasible subset; paired comparisons use joint successes.

\noindent\emph{Efficiency.} Final iTIMO releases replaced 283 DeepSeek single-disruption provider-balance errors via targeted reruns and 43 Gemini compound-disruption execution-error records through official-endpoint reruns. Metrics describe retained final outputs; discarded attempts are excluded from consumption aggregation. Where records are available, mean calls and tokens cover all retained cases, including failures. Qwen iTIMO figures use reported summaries. End-to-end latency is summarized in seconds by mean, median, and 95th percentile (P95), linearly interpolated where raw records are available. IPyHOPPER's zero calls/tokens mean no LLM inference, not zero computation.

\noindent\textbf{Implementation settings.}
LLM--Z3 and the iTIMO adapter use Gemini 3.8 Flash, DeepSeek V4 Flash, and Qwen 3.8 Flash service aliases. LLM--Z3 uses one call per case, temperature 0, a 2,048-token output limit, no retry, and disabled thinking for DeepSeek/Qwen. It checks up to 25 resource combinations without a similarity objective. IPyHOPPER uses no LLM, scorer-free disrupted-slot candidates, a 300-second limit, and no retry. The iTIMO adapter permits two planning rounds for one disruption and $k+1$ for $k$ simultaneous disruptions; format correction may add calls. Documented Gemini/DeepSeek settings use temperature 0 and a 2,048-token output limit. Qwen iTIMO adapter raw records, sampling settings, and P95 interpolation details are unavailable.

\subsection{Single-Disruption Evaluation}
\label{sec:single-results}

\noindent\textbf{Design.}
The single-disruption experiment evaluates isolated resource failures using 375 feasible repair cases and 125 infeasible cases, with effectiveness, stability, and efficiency assessed jointly.

\begin{table}[!htbp]
  \centering
  \footnotesize
  \setlength{\tabcolsep}{2pt}
  \caption{Single-disruption effectiveness and plan stability. Stability metrics use successful feasible repairs.}
  \label{tab:single-results}
  \begin{tabular}{llrrrrr}
    \toprule
    Method & LLM & \shortstack{Overall\\success (\%)} & \shortstack{Feasible\\repair (\%)} & \shortstack{Correct\\refusal (\%)} & \shortstack{Mean\\edits} & \shortstack{Preserved\\slots (\%)} \\
    \midrule
    LLM--Z3       & \shortstack[l]{Gemini 3.8\\Flash}   & 97.0 & 96.00 & 100.0 & 6.628 & 65.35 \\
    LLM--Z3       & \shortstack[l]{DeepSeek V4\\Flash} & 84.6 & 90.13 &  68.0 & 6.618 & 65.25 \\
    LLM--Z3       & \shortstack[l]{Qwen 3.8\\Flash}     & 91.0 & 88.53 &  98.4 & 6.572 & 65.36 \\
    IPyHOPPER     & --       & 96.8 & 95.73 & 100.0 & 1.000 & 92.67 \\
    iTIMO adapter & \shortstack[l]{Gemini 3.8\\Flash}   & 87.8 & 83.73 & 100.0 & 1.003 & 92.59 \\
    iTIMO adapter & \shortstack[l]{DeepSeek V4\\Flash} & 88.6 & 84.80 & 100.0 & 1.000 & 92.33 \\
    iTIMO adapter & \shortstack[l]{Qwen 3.8\\Flash}     & 67.0 & 73.33 &  48.0 & 1.000 & 91.95 \\
    \bottomrule
  \end{tabular}
\end{table}

\noindent\textbf{Effectiveness and plan stability.}
Table~\ref{tab:single-results} gives effectiveness in its first three metric columns and stability, conditional on successful feasible repairs, in its final two.
LLM--Z3 with Gemini achieved 485/500 overall successes (97.0\%) versus IPyHOPPER's 484/500 (96.8\%); this one-case difference alone supports no meaningful ranking. With DeepSeek, LLM--Z3 repaired 338/375 cases and correctly refused 85/125 conflicts, whereas the iTIMO adapter repaired 318/375 and refused all 125. Overall success therefore reversed their feasible-repair ordering.

Full replanning averaged about 6.6 edits and 65\% preservation, versus one edit and 92\% for successful IPyHOPPER and iTIMO adapter repairs. On the 346 jointly successful Gemini LLM--Z3/IPyHOPPER cases, respective means were 6.809 versus 1.000 edits and 64.70\% versus 92.68\% preservation. Restricting to joint successes removes differences from the methods succeeding on different case subsets; the comparison remains conditional on both returning valid repairs.

\begin{table}[!t]
  \centering
  \footnotesize
  \setlength{\tabcolsep}{3pt}
  \caption{Single-disruption efficiency over 500 cases. Calls and tokens are per-case means; latency is in seconds.}
  \label{tab:single-runtime}
  \begin{tabular}{llrrrrr}
    \toprule
    Method & LLM & Calls & Tokens & \shortstack{Mean\\(s)} & \shortstack{Median\\(s)} & \shortstack{P95\\(s)} \\
    \midrule
    LLM--Z3       & Gemini 3.8 Flash   & 1.000 & 1,369  & 5.10  & 3.89  & 9.08 \\
    LLM--Z3       & DeepSeek V4 Flash & 1.000 & 938    & 3.25  & 1.96  & 4.83 \\
    LLM--Z3       & Qwen 3.8 Flash     & 1.000 & 999    & 4.48  & 3.31  & 6.53 \\
    IPyHOPPER     & --                 & 0     & 0      & 2.05  & 1.98  & 2.57 \\
    iTIMO adapter & Gemini 3.8 Flash   & 1.244 & 23,469 & 14.21 & 13.00 & 29.72 \\
    iTIMO adapter & DeepSeek V4 Flash & 1.132 & 19,486 & 8.41  & 8.27  & 16.35 \\
    iTIMO adapter & Qwen 3.8 Flash     & 1.430 & 24,205 & 11.88 & 17.24 & 23.55 \\
    \bottomrule
  \end{tabular}
\end{table}

\noindent\textbf{Efficiency.}
Table~\ref{tab:single-runtime} reports per-case inference consumption and latency, subject to the record-availability qualification above.
LLM--Z3 uses one call and 938--1,369 tokens per case. The iTIMO adapter averaged 1.132--1.430 calls and 19,486--24,205 tokens despite approximately one logical edit per successful repair. IPyHOPPER required no LLM inference and recorded a 1.98-second median. Few edits do not imply low inference consumption; latency describes these implementations and environments, not intrinsic algorithm speed.

\noindent\textbf{Summary.}
LLM--Z3 with Gemini and IPyHOPPER achieved nearly identical overall success. Successful hierarchical and local repairs preserved more accepted commitments than full replanning, while computational costs differed substantially across strategies.

\subsection{Compound-Disruption Evaluation}
\label{sec:compound-results}

\noindent\textbf{Design.}
The 200 feasible compound cases comprise 50 each of AB, AC, BC, and ABC, representing simultaneous flight--hotel, flight--attraction, hotel--attraction, and flight--hotel--attraction disruptions; refusal is therefore inapplicable.

\begin{table}[!htbp]
  \centering
  \footnotesize
  \setlength{\tabcolsep}{2pt}
  \caption{Compound-disruption effectiveness and plan stability over 200 feasible cases. Stability metrics use successful repairs.}
  \label{tab:multi-results}
  \begin{tabular}{llrrrrrrr}
    \toprule
    Method & LLM & \shortstack{Overall\\success (\%)} & AB & AC & BC & ABC & \shortstack{Mean\\edits} & \shortstack{Preserved\\slots (\%)} \\
    \midrule
    LLM--Z3       & Gemini 3.8 Flash   & 93.0 & 47 & 44 & 49 & 46 & 9.065 & 57.25 \\
    LLM--Z3       & DeepSeek V4 Flash & 85.5 & 41 & 42 & 48 & 40 & 9.053 & 56.96 \\
    LLM--Z3       & Qwen 3.8 Flash     & 84.0 & 41 & 41 & 46 & 40 & 8.982 & 57.06 \\
    IPyHOPPER     & --                 & 89.0 & 47 & 41 & 49 & 41 & 2.230 & 86.73 \\
    iTIMO adapter & Gemini 3.8 Flash   & 67.0 & 46 & 36 & 33 & 19 & 2.142 & 86.16 \\
    iTIMO adapter & DeepSeek V4 Flash & 75.5 & 37 & 46 & 36 & 32 & 2.212 & 86.67 \\
    iTIMO adapter & Qwen 3.8 Flash     & 67.0 & 31 & 37 & 42 & 24 & 2.179 & 86.35 \\
    \bottomrule
  \end{tabular}
\end{table}
\noindent\textbf{Effectiveness and plan stability.}
Table~\ref{tab:multi-results} reports overall success over 200 cases, combination-level successful counts out of 50, and stability on successful repairs.
LLM--Z3 with Gemini repaired 186/200 cases (93.0\%), the highest observed success, followed by IPyHOPPER at 178/200 (89.0\%). The iTIMO adapter repaired 134, 151, and 134 cases with Gemini, DeepSeek, and Qwen. IPyHOPPER and LLM--Z3 with Gemini both repaired 47/50 AB and 49/50 BC cases; the iTIMO adapter with Gemini repaired 46/50 AB but 19/50 ABC cases. These differences do not establish a causal effect of disruption count because the groups use different itineraries.

Full replanning averaged about 9.0 edits and 57\% preservation, versus 2.1--2.2 edits and 86\%--87\% for hierarchical/local repair. On 167 jointly successful Gemini LLM--Z3/IPyHOPPER cases, respective means were 9.132 versus 2.222 edits and 57.28\% versus 86.77\% preservation. The paired comparison fixes the successful case subset and does not describe stability when either method fails.

\begin{table}[!htbp]
  \centering
  \footnotesize
  \setlength{\tabcolsep}{3pt}
  \caption{Compound-disruption efficiency over 200 cases. Calls and tokens are per-case means; latency is in seconds.}
  \label{tab:compound-runtime}
  \begin{tabular}{llrrrrr}
    \toprule
    Method & LLM & Calls & Tokens & \shortstack{Mean\\(s)} & \shortstack{Median\\(s)} & \shortstack{P95\\(s)} \\
    \midrule
    LLM--Z3       & Gemini 3.8 Flash   & 1.000 & 1,594  & 6.49  & 4.49  & 8.33 \\
    LLM--Z3       & DeepSeek V4 Flash & 1.000 & 1,086  & 4.85  & 2.57  & 7.77 \\
    LLM--Z3       & Qwen 3.8 Flash     & 1.000 & 1,173  & 6.33  & 4.09  & 12.08 \\
    IPyHOPPER     & --                 & 0     & 0      & 20.23 & 2.32  & 300.00 \\
    iTIMO adapter & Gemini 3.8 Flash   & 2.905 & 67,631 & 31.62 & 30.81 & 52.29 \\
    iTIMO adapter & DeepSeek V4 Flash & 1.660 & 34,843 & 13.19 & 12.32 & 23.96 \\
    iTIMO adapter & Qwen 3.8 Flash     & 1.845 & 39,182 & 19.29 & 16.79 & 38.76 \\
    \bottomrule
  \end{tabular}
\end{table}

\noindent\textbf{Efficiency.}
Table~\ref{tab:compound-runtime} reports per-case calls, tokens, and latency under the same provenance qualifications.
LLM--Z3 used one call and averaged 1,086--1,594 tokens; the iTIMO adapter averaged 1.660--2.905 calls and 34,843--67,631 tokens. IPyHOPPER required no LLM inference, but its 2.32-second median accompanied a 20.23-second mean, 300-second P95, and 12/200 timeouts. Median latency alone omits this tail; execution differences preclude an intrinsic speed ranking.

\noindent\textbf{Summary.}
LLM--Z3 with Gemini achieved the highest observed compound success. Hierarchical and local repairs preserved more accepted commitments, while IPyHOPPER showed a long latency tail and the iTIMO adapter higher token consumption.

\section{Practical Guidelines for Revision Strategy Selection}
\label{sec:guidelines}

No strategy dominates all evaluated dimensions. Table~\ref{tab:selection-guidelines} summarizes trade-offs within the evaluated setting.

\begin{table}[!htbp]
  \centering
  \footnotesize
  \setlength{\tabcolsep}{2pt}
  \caption{Strategy-selection guidance within the evaluated setting.}
  \label{tab:selection-guidelines}
  \begin{tabular}{@{}>{\raggedright\arraybackslash}p{0.25\textwidth}>{\raggedright\arraybackslash}p{0.30\textwidth}>{\raggedright\arraybackslash}p{0.40\textwidth}@{}}
    \toprule
    Priority & More attractive & Main trade-off / caution \\
    \midrule
    Feasibility recovery & LLM--Z3 / IPyHOPPER & Preservation / latency cost \\
    Preservation & IPyHOPPER / iTIMO adapter & Hierarchy / lower coverage \\
    Limited LLM inference & IPyHOPPER / LLM--Z3 & None / compact one-call inference \\
    Local semantic editing & iTIMO adapter & Bounded scope; locality not controlled \\
    \bottomrule
  \end{tabular}
\end{table}

\noindent\textbf{LLM--Z3: Feasibility Recovery, Compact Inference.}
LLM--Z3's observed coverage favors feasibility recovery when broad revisions are acceptable. Its larger changes are consistent with the absence of a preservation objective, and coverage remains configuration-dependent.

\noindent\textbf{IPyHOPPER: Strong Preservation, No LLM Inference.}
IPyHOPPER is attractive when preservation is prioritized and a usable hierarchy is available, requiring no LLM inference. Its compound-disruption latency tail matters under strict runtime requirements.

\noindent\textbf{iTIMO: Local Editing, Higher Inference Cost.}
The iTIMO adapter supports preservation-oriented local editing, but showed lower compound coverage and substantially higher token consumption in the evaluated configurations; locality was not independently varied.

These guidelines are limited to the evaluated implementations and synthetic resource-disruption setting. Repair scope is confounded with feedback, candidate restrictions, model routing, and budgets, so differences cannot be attributed solely to paradigms; no method is shown to return globally minimum-edit $P^*$. Refusal covers required-hotel conflicts only, while stability is conditioned on successful repairs and joint successes for paired comparisons. Compound groups use different itineraries and are not difficulty-matched. Runtime and token results depend on implementations, environments, and service endpoints. Exact historical equivalence for iTIMO cannot be established because its external knowledge-base/scorer checkout was not fingerprinted; Qwen iTIMO also lacks raw records, sampling settings, and P95 interpolation details.

\section{Conclusion}
\label{sec:conclusion}
We evaluated LLM--Z3 full replanning, IPyHOPPER hierarchical repair, and iTIMO-based local revision under single and compound resource disruptions. LLM--Z3 with Gemini achieved the highest observed compound success, while IPyHOPPER reached comparable single-disruption overall success with greater preservation. Hierarchical and local repairs made fewer edits than full replanning, while the strategies exhibited distinct computational costs. No strategy dominates all dimensions; selection should balance feasibility recovery, preservation, and computational cost.

%% file: fig_problem.tex
% Original TikZ artwork for the Introduction. Native size: 11.8 x 5.2 cm.
% Requires tikz and the arrows.meta library. No float or document preamble.
\begingroup%
\definecolor{prblue}{HTML}{48677E}%
\definecolor{prteal}{HTML}{50766F}%
\definecolor{prred}{HTML}{A65F53}%
\begin{tikzpicture}[x=1cm,y=1cm,
  font=\sffamily\fontsize{8}{9.4}\selectfont,text=black!90,
  draw=black!50,line width=.5pt,
  box/.style={draw,rounded corners=1.5pt,fill=white,align=center,inner sep=3pt},
  arrow/.style={-{Stealth[length=1.5mm,width=1.1mm]}},
  small/.style={font=\sffamily\fontsize{7.5}{8.7}\selectfont}]
\path[use as bounding box] (0,0) rectangle (11.8,5.2);
\node[anchor=west,font=\sffamily\bfseries\fontsize{9}{10.5}\selectfont]
  at (.12,4.98) {Itinerary Revision under Resource Disruptions};

% Accepted commitments precede disruptions.
\node[box,draw=prblue,fill=prblue!5,minimum width=2.05cm,minimum height=.74cm]
  (request) at (1.18,4.05) {Travel request $q$};
\draw[draw=prblue,rounded corners=2pt] (2.72,3.55) rectangle (11.65,4.65);
\node[anchor=west,small,text=prblue] at (2.9,4.43) {Accepted itinerary $P$};
\node[box,draw=prred,fill=prred!6,minimum width=2.1cm,minimum height=.45cm] (flight) at (4.13,3.95) {Flight};
\node[box,minimum width=2.1cm,minimum height=.45cm] (hotel) at (7.18,3.95) {Hotel};
\node[box,minimum width=2.1cm,minimum height=.45cm] (attraction) at (10.23,3.95) {Attraction};
\draw[arrow] (request.east)--(2.72,4.05);
\draw[arrow] (flight)--(hotel);
\draw[arrow] (hotel)--(attraction);
\draw[prred,line width=.9pt] (4.80,3.86)--(4.98,4.04) (4.80,4.04)--(4.98,3.86);

% Examples describe a set; the cross illustrates one affected component.
\node[box,draw=prred,fill=prred!4,minimum width=6.25cm,minimum height=.78cm]
  (delta) at (3.32,2.84) {\textbf{Resource disruptions $\Delta$}\\
  {\fontsize{7.5}{8.7}\selectfont Flight cancelled / hotel unavailable / attraction closed}};
\draw[arrow,prred,dashed] (4.13,3.23)--(flight.south);
\node[box,fill=black!3,minimum width=4.2cm,minimum height=.78cm]
  (environment) at (9.45,2.84) {Updated environment $\mathcal K'$\\$P$ may become infeasible};
\draw[arrow] (delta.east)--(environment.west);
\draw[arrow] (10.23,3.55)--(10.23,3.23);

% Revision is conditional; these are intended outcomes, not guaranteed success.
\node[small] at (1.18,3.43) {$q$ unchanged};
\node[box,draw=prblue,fill=prblue!6,minimum width=2.7cm,minimum height=.6cm]
  (revision) at (5.9,1.59) {\textbf{Revision}};
\draw[arrow] (environment.south)--(9.45,1.59)--(revision.east);
\node[small,fill=white,inner sep=2pt] at (9.45,2.03) {If $P$ is infeasible};
\node[box,draw=prteal,fill=prteal!5,minimum width=4.45cm,minimum height=.64cm]
  (revised) at (3.08,.43) {\textbf{Feasible revised itinerary $P'$}};
\node[box,draw=black!60,fill=black!3,minimum width=4.45cm,minimum height=.64cm]
  (diagnosis) at (8.72,.43) {\textbf{Infeasibility diagnosis $\bot$}};
\draw (revision.south)--(5.9,.97);
\draw[arrow] (5.9,.97)--(3.08,.97)--(revised.north);
\draw[arrow] (5.9,.97)--(8.72,.97)--(diagnosis.north);
\node[small,fill=white,inner sep=2pt] at (5.9,.43) {or};
\end{tikzpicture}%
\endgroup%

%% file: fig_evaluation.tex
% Original TikZ artwork for Experimental Evaluation. Native size: 11.8 x 7.3 cm.
% Requires tikz and the arrows.meta library. No float or document preamble.
\begingroup%
\definecolor{evblue}{HTML}{48677E}%
\definecolor{evteal}{HTML}{50766F}%
\definecolor{evorange}{HTML}{A37A49}%
\definecolor{evred}{HTML}{A65F53}%
\begin{tikzpicture}[x=1cm,y=1cm,
  font=\sffamily\fontsize{8}{9.4}\selectfont,text=black!90,
  draw=black!50,line width=.5pt,
  box/.style={draw,rounded corners=1.5pt,fill=white,align=center,inner sep=3pt},
  arrow/.style={-{Stealth[length=1.5mm,width=1.1mm]}},
  small/.style={font=\sffamily\fontsize{7.5}{8.7}\selectfont},
  method/.style={align=center,font=\sffamily\bfseries\fontsize{9}{10.5}\selectfont}]
\path[use as bounding box] (0,0) rectangle (11.8,7.3);
\node[anchor=west,font=\sffamily\bfseries\fontsize{9}{10.5}\selectfont]
  at (.12,7.07) {Unified Evaluation of Representative Revision Strategies};
\node[box,fill=black!3,minimum width=7.8cm,minimum height=.66cm]
  (input) at (5.9,6.43) {\textbf{Common revision instance} $(q,P,\Delta,\mathcal K')$\\
  {\fontsize{7.5}{8.7}\selectfont Same fixed instance for all methods; method-specific representations}};
\node[small,align=center] at (2.98,5.66) {\textbf{500 single-disruption cases}\\375 feasible + 125 infeasible};
\node[small,align=center] at (8.82,5.66) {\textbf{200 compound-disruption cases}\\All 200 feasible};
\draw (input.south)--(5.9,5.22);
\draw (2.05,5.22)--(9.75,5.22);
\foreach \x in {2.05,5.9,9.75} {\draw[arrow] (\x,5.22)--(\x,4.97);}

% Three different scopes, not three identical internal pipelines.
\draw[box,draw=evblue,fill=evblue!4] (.25,1.83) rectangle (3.85,4.97);
\draw[box,draw=evteal,fill=evteal!4] (4.1,1.83) rectangle (7.7,4.97);
\draw[box,draw=evorange,fill=evorange!4] (7.95,1.83) rectangle (11.55,4.97);
\node[method] at (2.05,4.55) {LLM--Z3\\Full Replanning};
\node[method] at (5.9,4.55) {IPyHOPPER\\Hierarchical Repair};
\node[method] at (9.75,4.55) {iTIMO\\Local Revision};
\node[small,text=evblue] at (2.05,4.02) {GLOBAL};
\node[small,text=evteal] at (5.9,4.02) {ADAPTIVE HIERARCHICAL};
\node[small,text=evorange] at (9.75,4.02) {BOUNDED LOCAL};

% Full replanning solves anew without implying that every component changes.
\node[small] at (2.05,3.63) {Solve updated problem};
\draw[arrow,evblue] (2.05,3.45)--(2.05,3.21);
\node[box,draw=evblue,fill=evblue!12,minimum width=2.65cm,minimum height=.43cm,inner sep=1pt]
  at (2.05,2.97) {Complete plan $P'$};
\node[small,align=center] at (2.05,2.32) {Formal constraint solving\\No similarity objective on $P$};

% Retained tree on the left; crossed failed branch and a solid repair on right.
\node[small] at (5.9,3.63) {Reconstruct hierarchy from $P$};
\draw[evteal] (5.77,3.31)--(4.91,2.99) (4.91,2.99)--(4.55,2.68)
  (4.91,2.99)--(5.27,2.68) (5.77,3.31)--(6.66,2.99);
\draw[evred,dashed] (6.66,2.99)--(6.25,2.68);
\draw[evred,line width=.85pt] (6.17,2.60)--(6.33,2.76) (6.17,2.76)--(6.33,2.60);
\draw[evteal] (6.66,2.99)--(7.08,2.68);
\foreach \x/\y in {5.77/3.31,4.91/2.99,4.55/2.68,5.27/2.68,6.66/2.99} {
  \filldraw[draw=evteal,fill=white] (\x,\y) circle (.055);}
\filldraw[draw=evteal,fill=evteal!20] (7.08,2.68) rectangle (7.19,2.79);
\draw[arrow,evteal] (6.36,2.65) to[bend right=22] (6.98,2.65);
\draw[arrow,evteal] (7.28,3.10) to[bend right=25] (6.03,3.34);
\node[small,align=center] at (5.9,2.17) {HTN repair + backtracking\\Repair may expand upward};

% Direct local editing preserves the two unedited symbols.
\node[small] at (9.75,3.63) {Direct semantic edits to $P$};
\foreach \x/\t in {9.75/H,10.80/A} {
  \node[box,minimum width=.65cm,minimum height=.43cm,inner sep=1pt]
    at (\x,3.10) {$\t$};}
\node[box,draw=evorange,fill=evorange!14,minimum width=.65cm,minimum height=.43cm,inner sep=1pt]
  at (8.70,3.10) {$F'$};
\draw[arrow] (9.04,3.10)--(9.41,3.10);
\draw[arrow] (10.09,3.10)--(10.46,3.10);
\draw[evorange] (8.38,2.83)--(9.02,2.83);
\node[small,align=center] at (9.75,2.32) {ADD / REPLACE / DELETE\\Preserve untouched\\components};

% One common evaluator; outcome categories and stability metrics stay separate.
\foreach \x in {2.05,5.9,9.75} {\draw (\x,1.83)--(\x,1.58);}
\draw (2.05,1.58)--(9.75,1.58);
\draw[arrow] (5.9,1.58)--(5.9,1.31);
\draw[box,fill=black!3] (.25,.10) rectangle (11.55,1.31);
\node[font=\sffamily\bfseries\fontsize{8.5}{10}\selectfont] at (5.9,1.14) {Common Evaluation};
\draw[black!25] (4.02,.23)--(4.02,.83) (7.78,.23)--(7.78,.83);
\node at (2.14,.82) {\textbf{Effectiveness}};
\node[small,align=center] at (2.14,.40) {repair / correct refusal /\\overall success};
\node[align=center] at (5.9,.64) {\textbf{Plan Stability}\\{\fontsize{7.5}{8.7}\selectfont logical edits / preservation}};
\node[align=center] at (9.66,.64) {\textbf{Computational Cost}\\{\fontsize{7.5}{8.7}\selectfont calls / tokens / latency}};
\end{tikzpicture}%
\endgroup%